\documentclass{article}
\usepackage{spconf,amsmath,graphicx,hyperref}
\usepackage{balance}
\usepackage{orcidlink}
\usepackage{float}
\usepackage{amsmath,amsfonts}
\usepackage{algorithmic}
\usepackage{array}
\usepackage[caption=false,font=normalsize,labelfont=sf,textfont=sf]{subfig}
\usepackage{textcomp}
\usepackage{stfloats}
\usepackage{url}
\usepackage{verbatim}
\usepackage{graphicx}

\usepackage[utf8]{inputenc} %
\usepackage[T1]{fontenc}    %
\usepackage{hyperref}       %
\usepackage{url}            %
\usepackage{booktabs}       %
\usepackage{amsfonts}       %
\usepackage[ruled,vlined]{algorithm2e} %
\usepackage{graphicx}       %
\usepackage{nicefrac}       %
\usepackage{microtype}      %
\usepackage{xcolor}         %
\usepackage{multirow}
\usepackage{amssymb}
\usepackage{amsmath}

\usepackage{wrapfig} %
\usepackage{booktabs}  %
\usepackage{lipsum}    %

\title{Embracing Aleatoric Uncertainty in Medical Multimodal Learning with Missing Modalities}

  \name{Linxiao Gong$^1$, Yang Liu$^2$, Lianlong Sun$^3$, Yulai Bi$^4$, Jing Liu$^{5,6}$, Xiaoguang Zhu$^{7*}$\thanks{$^*$Corresponding author.}}
  \address{$^1$HKUST (GZ) \quad $^2$Tongji University \quad $^3$University of Rochester \quad $^4$Meta \\ \quad $^5$Fudan University \quad $^6$The University of British Columbia \quad $^7$University of California, Davis}
\begin{document}
\maketitle
\begin{abstract}
Medical multimodal learning faces significant challenges with missing modalities prevalent in clinical practice. Existing approaches assume equal contribution of modality and random missing patterns, neglecting inherent uncertainty in medical data acquisition. In this regard, we propose the \textbf{A}leatoric \textbf{U}ncertainty \textbf{M}odeling (\textbf{AUM}) that explicitly quantifies unimodal aleatoric uncertainty to address missing modalities. Specifically, AUM models each unimodal representation as a multivariate Gaussian distribution to capture aleatoric uncertainty and enable principled modality reliability quantification. To adaptively aggregate captured information, we develop a dynamic message-passing mechanism within a bipartite patient-modality graph using uncertainty-aware aggregation mechanism. Through this process, missing modalities are naturally accommodated, while more reliable information from available modalities is dynamically emphasized to guide representation generation. Our AUM framework achieves an improvement of 2.26\% AUC-ROC on MIMIC-IV mortality prediction and  2.17\% gain on eICU, outperforming existing state-of-the-art approaches.

\end{abstract}


\section{Introduction}
\label{sec:intro}

Multimodal data provides comprehensive information for clinical 
support in modern medical practice \cite{maltesen2020longitudinal,wagner2020ptb,lu2025impact,liu2025privacy,su2025large,liu2025multimodal}. To integrate these data, multimodal representation learning leverages deep learning techniques to fuse heterogeneous data sources and achieve more accurate disease prediction and diagnosis \cite{scholkopf2021toward,liu2025crcl,liu2025multimodal}. However, complete multimodal data is often difficult to obtain due to various factors, including cost constraints, technical limitations, and patient compliance issues \cite{lin2023dynamically,ford2000non,pan2021disease}. 

To address the challenge of missing modalities, researchers have proposed solutions that can be categorized into two main approaches. Modality imputation methods employ deep generative models to reconstruct missing modality information \cite{ngiam2011multimodal,ma2021smil,tran2017missing}. However, these approaches rely on strong prior assumptions to learn mappings from low-dimensional latent representations to high-dimensional raw data spaces. %
In contrast, some direct prediction methods bypass the modality reconstruction step and directly utilize observable incomplete data for prediction with deep learning models \cite{chen2020hgmf,wu2024multimodal,zhang2022m3care, LIU2022109697,zhu2025causal}. %
While this approach avoids the accumulation of reconstruction errors, it still faces significant limitations. Both categories of methods typically assume the reconstruction of missing modalities uniform weighting of contributions from available modalities, and neglect %
the inherent data uncertainty in the data acquisition and reconstruction process \cite{you2020handling,ramachandram2017deep}.
However, these assumptions and data uncertainty highlight the need to consider differential modality contributions to clinical predictions \cite{cornelisz2020addressing,zhang2025uncertainty}.

\begin{figure}[t]
    \centering 
    \includegraphics[width=0.5\textwidth]{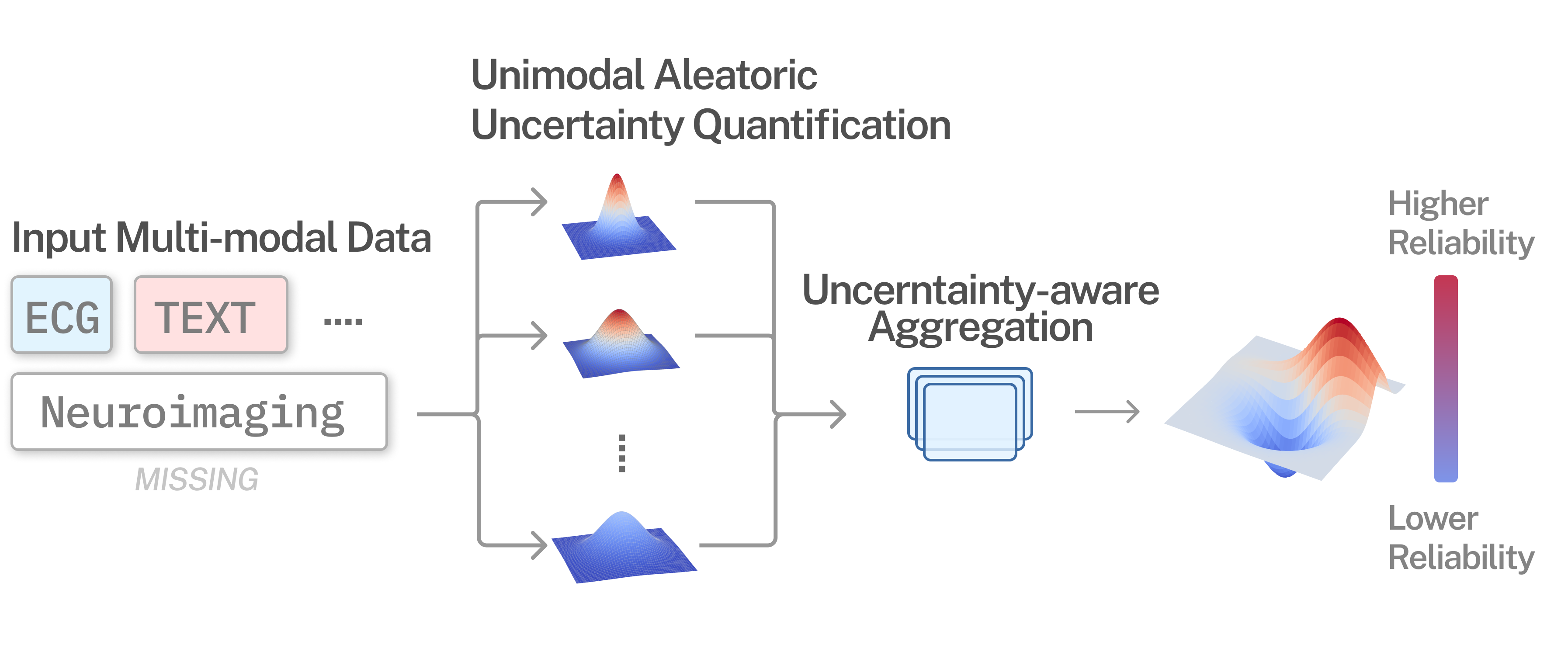}
    \vspace{-10pt}
    \caption{Overview of the proposed AUM framework.}
    \label{fig: Overview}
     \vspace{-5pt}
\end{figure}

Clinical evidence suggests that medical data exhibits high aleatoric uncertainty across different patients and conditions \cite{zhang2025uncertainty}. For example, elderly patients with similar demographics may exhibit comparable brain atrophy on MRI imaging, yet belong to different diagnostic categories: some have Alzheimer's disease, while others experience normal aging.
Furthermore, systematic patterns in data availability can compound these challenges \cite{castro2020causality}. For example, missing of neuroimaging data for patients with advanced dementia in ADNI datasets introduces significant data uncertainty, leading to poor prediction for the patient population.

To address these challenges, this paper proposes an uncertainty-aware multimodal relationship modeling approach, named \textbf{A}leatoric \textbf{U}ncertainty \textbf{M}odeling (\textbf{AUM}). Our core insight is to explicitly quantify the aleatoric uncertainty of each modality and dynamically adjust the patient representation learning process based on this uncertainty. 
Specifically, we assign uncertainty measures to each modality according to its inherent noise levels and reliability. Moreover, AUM naturally handles missing modalities through high-uncertainty initialization, eliminating explicit data imputation. Fig.~\ref{fig: Overview} presents the overall architecture of AUM framework.

\section{Methodology}

\subsection{Preliminary and Baseline}

In this paper, we employ MUSE \cite{wu2024multimodal} as our baseline. MUSE addresses the challenge of missing modalities through a bipartite patient-modality graph representation combined with mutually consistent contrastive learning.

The core architecture constructs an undirected bipartite graph $\mathcal{G} = (\mathcal{V}, \mathcal{E})$ where the node set consists of patient nodes and modality nodes. Edge features are initialized using modality-specific encoders for modality input $x_m$: $e_{u_p v_m} = \text{Encoder}_m(x_m)$. Patient representations are learned through multi-layer Graph Neural Network message passing.

MUSE introduces a mutual-consistent contrastive learning mechanism by generating an augmented graph through random edge dropout and processing both graphs via a Siamese GNN architecture. The framework optimizes a combination of unsupervised contrastive, supervised contrastive, and classification losses. Our approach extends this foundation by explicitly modeling the reliability and informativeness of each modality through uncertainty quantification.

\subsection{Unimodal Aleatoric Uncertainty Quantification}

To learn robust unimodal representations in medical multimodal data with inherent aleatoric uncertainty, we model each unimodal sample as a multivariate Gaussian distribution. For a given multimodal sample with modality $m \in \mathcal{M}$, where $\mathcal{M}$ represents the set of all modalities, we first employ modality-specific encoders to extract preliminary feature representations. Subsequently, we deploy two separate multilayer perceptrons to learn the distributional parameters:
\begin{equation}
\mu_i^m = f_{\theta_1^m}(\mathbf{x}_i^m) \quad \sigma_i^m = f_{\theta_2^m}(\mathbf{x}_i^m),
\end{equation}
where $f_{\theta_1^m}(\cdot)$ and $f_{\theta_2^m}(\cdot)$ represent two different fully connected layers for mean and variance estimation respectively, $\mathbf{x}_i^m$ denotes the input feature of modality $m$ for patient $i$, and $\mu_i^m \in \mathbb{R}^d$, $\sigma_i^m \in \mathbb{R}^d$ are the learned mean and variance vectors with dimension $d$.

The unimodal representation $\mathbf{z}_i^m$ in the latent space is then characterized as a multivariate Gaussian distribution:
\begin{equation}
p(\mathbf{z}_i^m | \mathbf{x}_i^m) \sim \mathcal{N}(\mu_i^m, (\sigma_i^m)^2 \mathbf{I}),
\end{equation}
where $\mathbf{I}$ is the identity matrix. The variance $(\sigma_i^m)^2$ quantifies the aleatoric uncertainty inherent in modality $m$, while the mean $\mu_i^m$ represents the corresponding feature representation.

\subsection{Dynamic Message Passing}

To effectively handle uncertainty in medical multimodal data, we represent each node and edge in the bipartite graph as a Gaussian distribution. Specifically, for the $l$-th layer, node embeddings are represented as $\mathbf{h}_i^{(l)} = (\mu_i^{(l)}, \sigma_i^{(l)})$ and edge embeddings as $\mathbf{e}_{ji}^{(l)} = (\mu_{ji}^{(l)}, \sigma_{ji}^{(l)})$, where each represents a Gaussian distribution $\mathcal{N}(\mu, \sigma^2 I)$. We compute the distributional parameters through linear transformations with ReLU activation for means and softplus functions for variances, ensuring the updated edge embeddings contain complete distributional information.

Based on the constructed messages, we design an uncertainty-aware attention mechanism. The attention weight $\alpha_{ji}^{(l)}$ is defined based on the inverse uncertainty of the message, such that messages with low uncertainty receive greater weights:
\begin{equation}
\alpha_{ji}^{(l)} = \frac{\exp\left( \tfrac{\vartheta}{\bar\sigma_{ji}^{(l)}} \right)}{\sum\limits_{k \in \mathcal{N}(i)} \exp\left( \tfrac{\vartheta}{\bar\sigma_{ki}^{(l)}} \right)},
\end{equation}
where %
$\vartheta$ is a temperature parameter.

In the node update phase, we employ a Gaussian aggregation strategy that handles mean and variance separately. For mean aggregation, we concatenate the node's historical information with weighted messages from neighbors:
\begin{equation}
\mu_i^{(l)} = \mathbf{U}^{(l)}\left[ \mu_i^{(l-1)} \Big\| \sum\limits_{j \in \mathcal{N}(i)} \alpha_{ji}^{(l)} \mu_{ji}^{(l)} \right].
\end{equation}
Variance aggregation follows the combination rule for independent Gaussian variables, considering the effect of attention weights on uncertainty propagation:
\begin{equation}
\left(\sigma_i^{(l)}\right)^2 = \left(\sigma_i^{(l-1)}\right)^2 + \sum\limits_{j \in \mathcal{N}(i)} \left(\alpha_{ji}^{(l)}\right)^2 \cdot \left(\sigma_{ji}^{(l)}\right)^2.
\label{eq:variance_aggregation}
\end{equation}
This update approximates the composition of independent Gaussian messages under uncertainty-aware attention, which effectively ensures the rationality of uncertainty propagation.

\begin{table*}[h]
  \centering
  \caption{Results on ICU datasets. A dagger ($\dagger$) indicates the standard deviation is greater than 0.02. An asterisk (*) indicates that the AUM-based models achieve a significant improvement over the baselines, with a p-value smaller than 0.05.}
  \scalebox{0.95}{
  \renewcommand{\arraystretch}{0.8}
  \setlength{\tabcolsep}{4pt}
  \begin{tabular}{lcccccccc}
      \toprule
      \multirow{3.7}{*}{\centering \textbf{Method}} & \multicolumn{4}{c}{\textbf{MIMIC-IV}} & \multicolumn{4}{c}{\textbf{eICU}} \\
      \cmidrule(lr){2-5} \cmidrule(lr){6-9}
      & \multicolumn{2}{c}{\textbf{Mortality}} & \multicolumn{2}{c}{\textbf{Readmission}} & \multicolumn{2}{c}{\textbf{Mortality}} & \multicolumn{2}{c}{\textbf{Readmission}} \\
      \cmidrule(lr){2-3} \cmidrule(lr){4-5} \cmidrule(lr){6-7} \cmidrule(lr){8-9}
      & \textbf{AUC-ROC} & \textbf{AUC-PRC} & \textbf{AUC-ROC} & \textbf{AUC-PRC} & \textbf{AUC-ROC} & \textbf{AUC-PRC} & \textbf{AUC-ROC} & \textbf{AUC-PRC} \\ 
      \midrule
      CM-AE \cite{ngiam2011multimodal}   & 0.8530$^\dagger$ & 0.4351$^\dagger$ & 0.6817$^\dagger$ & 0.4324$^\dagger$ & 0.8624 & 0.3902 & 0.7462$^\dagger$ & 0.4338$^\dagger$ \\
      SMIL \cite{ma2021smil}   & 0.8607 & 0.4438 & 0.6894$^\dagger$ & 0.4368$^\dagger$ & 0.8711 & 0.4066 & 0.7506 & 0.4447 \\
      MT \cite{ma2022multimodal}    & 0.8739 & 0.4452 & 0.6901 & 0.4375 & 0.8882 & 0.4109 & 0.7635 & 0.4500 \\
      GRAPE \cite{you2020handling}   & 0.8837 & 0.4584$^\dagger$ & 0.7085 & 0.4551 & 0.8903 & 0.4137 & 0.7663 & 0.4501 \\
      HGMF \cite{chen2020hgmf}   & 0.8710 & 0.4433 & 0.7005$^\dagger$ & 0.4421 & 0.8878 & 0.4104 & 0.7604 & 0.4496$^\dagger$ \\
      M3Care \cite{zhang2022m3care}  & 0.8896$^\dagger$ & 0.4603$^\dagger$ & 0.7067 & 0.4532 & 0.8964 & 0.4155 & 0.7598$^\dagger$ & 0.4430 \\
      MUSE \cite{wu2024multimodal}  & 0.9004 & 0.4735 & 0.7152 & 0.4670$^\dagger$ & 0.9017 & 0.4216 & 0.7709 & 0.4631 \\
      \midrule
      \textbf{GRAPE+AUM}   & 0.9085* & 0.4775* & 0.7197* & 0.4939* & 0.9212* & 0.4279* & 0.7821* & 0.4736* \\
      \textbf{MUSE+AUM}   & \textbf{0.9230*} & \textbf{0.4926*} & \textbf{0.7281*} & \textbf{0.5003*} & \textbf{0.9234*} & \textbf{0.4382*} & \textbf{0.7879*} & \textbf{0.4826*} \\
      \bottomrule
  \end{tabular}
  }
  \label{tab:icu}
  \vspace{-15pt}
\end{table*}

\subsection{Loss Function}

We augment the base loss function of MUSE with an additional Variational Information Bottleneck (VIB) term to constrain the uncertainty learning process \cite{gao2024embracing}. Specifically, we inherit three components from MUSE: unsupervised contrastive loss $\mathcal{L}_\text{unsup}$ that encourages consistency between representations of the same patient in different modality configurations, supervised contrastive loss $\mathcal{L}_\text{sup}$ that enhances similarity among patients with identical labels, and standard classification loss $\mathcal{L}_\text{CE}$ between predicted $\hat{\mathbf{y}}_p$ with groundtruth $\mathbf{y}_p$. To regulate the learned uncertainty distributions and prevent overfitting, we incorporate the VIB mechanism, which restricts the patient representations to follow a regularized distribution. The complete loss function is as follows:
\begin{equation}
\resizebox{0.91\linewidth}{!}{$
\begin{aligned}
\mathcal{L}_\text{total} = & \mathcal{L}_\text{unsup}(\mathbf{Z}, \mathbf{Z}') 
+ \mathcal{L}_\text{sup}(\mathbf{Z}, \mathbf{Y}) \\
&+ \textstyle\sum_{p=1}^N \mathcal{L}_\text{CE}(\hat{\mathbf{y}}_p, \mathbf{y}_p) 
+ \textstyle\sum_{p=1}^N \mathcal{L}_\text{VIB}(\boldsymbol{\mu}, \boldsymbol{\sigma}),
\end{aligned}$}
\end{equation}
where $\mathbf{Z} = \{\mu_p\}_{p=1}^N$ and $\mathbf{Z}' = \{\mu_p'\}_{p=1}^N$ are the mean components of patient representations from original and augmented graphs, and the VIB term is defined as:
\begin{equation}
  \resizebox{0.9\linewidth}{!}{$
\mathcal{L}_\text{VIB}(\boldsymbol{\mu}, \boldsymbol{\sigma}) = \sum_{p=1}^N \text{KL}\left(\mathcal{N}(\mu_p, (\sigma_p)^2 \mathbf{I}) \| \mathcal{N}(\mathbf{0}, \mathbf{I})\right)
$}. 
\end{equation}

\section{Experiments and results}

\subsection{Setup}
\textbf{Implementation Details.} We train the model for 100 epochs on an NVIDIA A100 GPU. To ensure stable convergence of the uncertainty quantification components, we adopt a two-stage training strategy. In the first 20 epochs, we keep the variance layers frozen and only train the mean prediction components. For the remaining 80 epochs, all parameters are jointly optimized. We select the best-performing model based on the validation set's performance.

\noindent \textbf{Datasets.}
We evaluated our proposed method on two widely used critical care databases: MIMIC-IV and eICU. MIMIC-IV \cite{johnson2023mimic} contains de-identified electronic health records from Beth Israel Deaconess Medical Center, including over 65,000 ICU admissions and 200,000 emergency department visits collected between 2008 and 2019. eICU \cite{pollard2018eicu} is a multicenter database covering more than 200,000 ICU admissions from 335 units in 208 hospitals across the United States (2014-2015). Both datasets are extensively validated in the medical informatics community, and naturally exhibit the missing modality patterns that our approach aims to address.

\begin{figure}[t]
  \centering
  \includegraphics[width=0.45\textwidth]{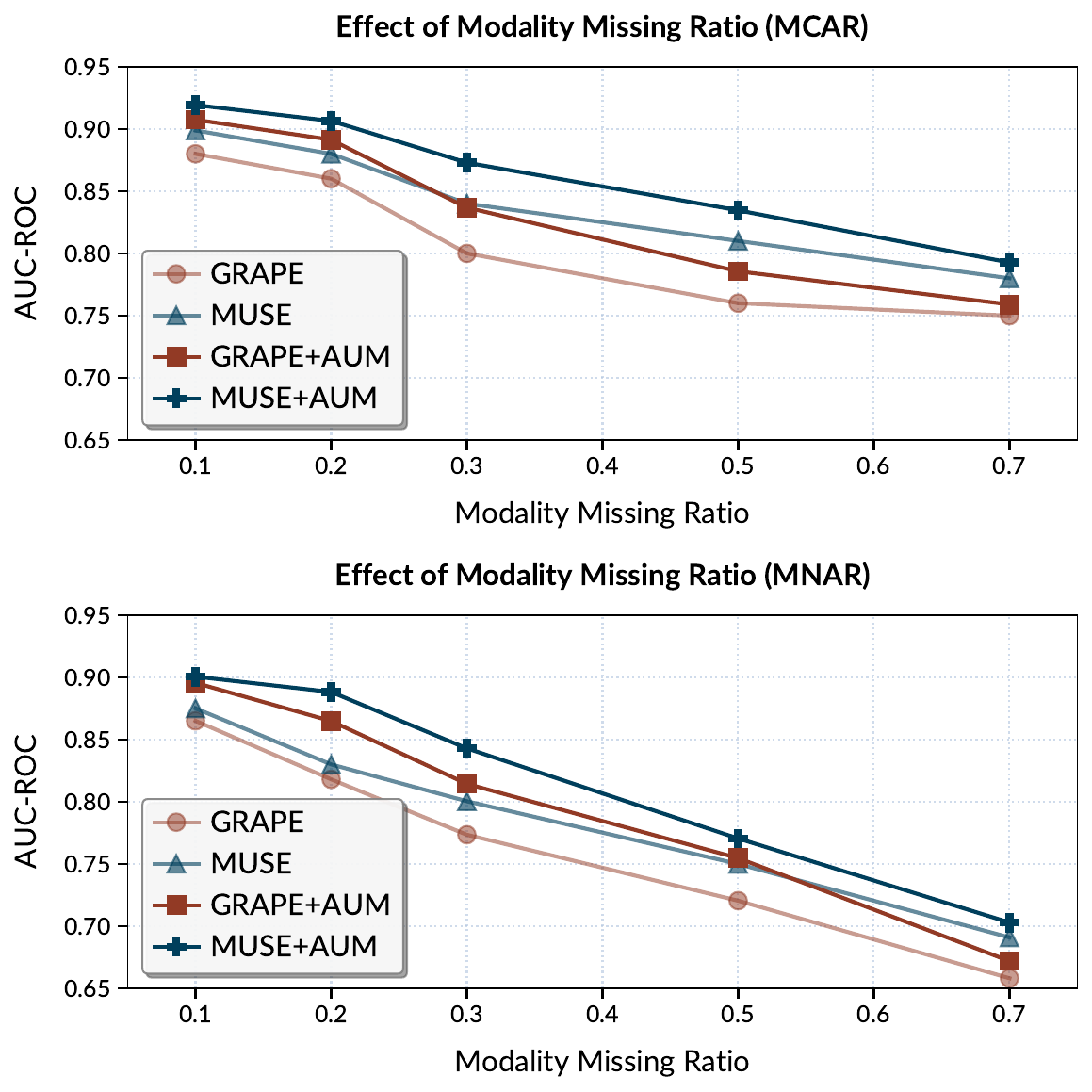}
  \caption{Performance comparison under different missing ratios.}
  \label{fig: missing ratio experiments}
  \vspace{-15pt}
\end{figure}

\subsection{Quantitative Comparisons}
\label{mainresults}

We conduct a systematic evaluation of the proposed method on the MIMIC-IV \cite{johnson2023mimic} and eICU \cite{pollard2018eicu} datasets.
Table~\ref{tab:icu} presents the experimental results in the MIMIC-IV and eICU datasets for the prediction tasks of mortality and readmission. The AUM-enhanced models consistently outperform all baseline methods with statistical significance  ($p < 0.05$). Among the baseline methods, MUSE achieves the best performance, followed by M3Care and GRAPE. %
Our method demonstrates substantial improvements: MUSE+AUM achieves an improvement of 2.26\% AUC-ROC for MIMIC-IV and 2.17\% for mortality from the eICU.

\subsection{Performance Under Different Missing Ratios}

To further evaluate the robustness of our proposed method under varying degrees of modality incompleteness, we conduct experiments across different missing ratios ranging from 0.1 to 0.7. Fig.~\ref{fig: missing ratio experiments} illustrates the performance comparison under both Missing Completely At Random (MCAR) and Missing Not At Random (MNAR) scenarios. %
Our methods demonstrate robust performance up to 0.5 missing ratio under MCAR and 0.3 under more challenging MNAR scenario, obtaining an AUC-ROC of approximately 0.85.
Notably, MUSE+AUM consistently outperforms all baseline methods across both missingness patterns, %
demonstrating the effectiveness of uncertainty-aware modeling in handling incomplete multimodal data.

\begin{table*}[htbp]
\centering
\caption{Ablation study on the influence of each module and variant.}
\label{tab:ablation1}
\setlength{\tabcolsep}{2.6pt}
\renewcommand{\arraystretch}{0.9}
\resizebox{0.75\textwidth}{!}{
\begin{tabular}{llcccc}
\toprule
\multirow{2}{*}{\textbf{ID}} & \multirow{2}{*}{\textbf{Method}} & \multicolumn{2}{c}{\textbf{MIMIC-IV}} & \multicolumn{2}{c}{\textbf{eICU}}  \\
\cmidrule(lr){3-4} \cmidrule(lr){5-6} 
 &  & \textbf{Mortality} & \textbf{Readmission} & \textbf{Mortality} & \textbf{Readmission} \\
 \midrule
- & MUSE \cite{wu2024multimodal}  & 0.9004 & 0.7152 & 0.9017 & 0.7709 \\
\midrule
A1.1 & MUSE+AUM w/o VIB & 0.9218 & 0.7243 & 0.9144 & 0.7749  \\
A1.2 & MUSE+AUM w/ learned variance in Eq.~\eqref{eq:variance_aggregation} & 0.9045 & 0.7139 & 0.9124 & 0.7778  \\
A1.3 & MUSE+AUM w/o unimodal uncertainty modeling & 0.9187 & 0.7246 & 0.9097 & 0.7858  \\
\midrule
- & MUSE+AUM & 0.9230 & 0.7281 & 0.9234 & 0.7879 \\
\bottomrule
\end{tabular}}
\vspace{-15pt}
\end{table*}

\begin{table}[htbp]
\centering
\caption{Performance comparison on the noisy dataset.}
\label{tab:ablation2}
\setlength{\tabcolsep}{2.6pt}
\renewcommand{\arraystretch}{1}
\resizebox{\columnwidth}{!}{%
\begin{tabular}{llcccc}
\toprule
\multirow{2}{*}{\textbf{$\epsilon$}} & \multirow{2}{*}{\textbf{Method}} & \multicolumn{2}{c}{\textbf{MIMIC-IV}} & \multicolumn{2}{c}{\textbf{eICU}}  \\
\cmidrule(lr){3-4} \cmidrule(lr){5-6} 
 &  & \textbf{Mortality} & \textbf{Readmission} & \textbf{Mortality} & \textbf{Readmission} \\
\midrule
\multirow{4}{*}{0} & MT & 0.8739 & 0.6901 & 0.8882 & 0.7635  \\
 & GRAPE & 0.8837 & 0.7085 & 0.8903 & 0.7663  \\
 & MUSE & 0.9004 & 0.7152 & 0.9017 & 0.7709 \\
 & MUSE+AUM & 0.9230 & 0.7281 & 0.9234 & 0.7879 \\
\midrule
\multirow{4}{*}{0.1} & MT & 0.8223 & 0.6299 & 0.8145 & 0.7158  \\
 & GRAPE & 0.8568 & 0.6577 & 0.8504 & 0.7322  \\
 & MUSE & 0.8694 & 0.6823 & 0.8746 & 0.7420  \\
 & MUSE+AUM & 0.8934 & 0.7073 & 0.8981 & 0.7554 \\
\midrule
\multirow{4}{*}{0.3} & MT & 0.7556 & 0.5856 & 0.7625 & 0.6638  \\
 & GRAPE & 0.7768 & 0.6014 & 0.7865 & 0.6911  \\
 & MUSE & 0.8012 & 0.6354 & 0.8187 & 0.7236  \\
 & MUSE+AUM & 0.8245 & 0.6477 & 0.8404 & 0.7324 \\
\bottomrule
\end{tabular}
}
\vspace{-15pt}
\end{table}

\subsection{Noise Robustness Analysis}

To evaluate the robustness of our proposed method under noisy environments, we conduct noise injection experiments on the MIMIC-IV and eICU datasets. Specifically, we add Gaussian noise of varying intensities to the original data, with noise levels $\epsilon$ set to 0, 0.1, and 0.3, where $\epsilon$ represents the ratio of noise standard deviation to signal standard deviation.

As shown in Table~\ref{tab:ablation2}, the performance of all methods degrades as the noise level increases. Under clean data conditions %
, MUSE+AUM achieves an AUC-ROC of 0.9230 on the MIMIC-IV mortality prediction task, significantly outperforming MUSE. Under moderate noise conditions%
, %
MUSE+AUM maintains its leading advantage with an AUC-ROC of 0.8934, compared to MUSE's 0.8694, demonstrating substantial improvement. Under high noise conditions%
, the performance gap further widens, with MUSE+AUM achieving 0.8245 while MUSE only reaches 0.8012, indicating that our method exhibits stronger robustness in noisy environments.

\subsection{Ablation Study}

Table~\ref{tab:ablation1} presents ablation studies that evaluate the contribution of each component in our proposed framework. We analyze three key variants: removing the VIB component, using an MLP to directly learn node uncertainty instead of the theoretical variance aggregation in Eq.~\eqref{eq:variance_aggregation}, and removing unimodal uncertainty modeling.
The results demonstrate that each component contributes to the overall performance. A1.1 shows significant performance drops, with MIMIC-IV mortality AUC-ROC decreasing from 0.9230 to 0.9218, indicating the importance of variational information bottleneck for learning compact representations. A1.2 exhibits a greater performance degradation, decreasing from 0.9230 to 0.9045, validating our theoretical approach to uncertainty aggregation in Eq.~\eqref{eq:variance_aggregation}. A1.3 also shows consistent drops across all metrics, decreasing from 0.9230 to 0.9187, confirming the effectiveness of modeling uncertainty at the unimodal level. These results validate each proposed component of our framework for robust multimodal learning with missing modalities. Additionally, since our framework only incorporates MLPs for variance representation, the computational complexity increase is minimal.

\section{Conclusion}

This paper proposed AUM, an uncertainty-aware multimodal learning framework that addresses incomplete modalities in clinical data by explicitly quantifying unimodal aleatoric uncertainty. Our approach represents each modality as a multivariate Gaussian distribution, leveraging variance parameters to capture inherent data noise. We designed an uncertainty-aware dynamic graph-based message-passing mechanism that naturally handles missing modalities through high-uncertainty initialization, circumventing error accumulation in conventional reconstruction methods. Extensive experiments on MIMIC-IV and eICU datasets demonstrate significant performance improvements: Mortality prediction achieves AUC-ROC gains of 2.26\% and 2.17\%, respectively. %
This work establishes a more robust multimodal fusion paradigm for clinical decision support systems. Future work will focus on the mathematical modeling of uncertainty propagation and applications in interpretable diagnostics.

\newpage
\small
\bibliographystyle{IEEEbib}
\bibliography{strings,refs}

\end{document}